\documentclass[10pt,twocolumn,letterpaper]{article}
\usepackage[pagenumbers]{cvpr}
\usepackage{float}
\usepackage{microtype}
\definecolor{cvprblue}{rgb}{0.21,0.49,0.74}
\usepackage[breaklinks,colorlinks,allcolors=cvprblue]{hyperref}

\title{LightSEAD: Lightweight SE(3)-Aware Keypoint Detector through Selective Knowledge Distillation}
\author{Mayowa Adebambo\textsuperscript{1}\quad
Alexander D. Rast\textsuperscript{2}\quad
Andrew Bradley\textsuperscript{1}\\[4pt]
\small\textsuperscript{1}Autonomous Driving and Intelligent Transport Group (ADIT), Oxford Brookes University\\
\small\textsuperscript{2}Artificial Intelligence, Data Analysis and Systems (AIDAS) Institute\\[3pt]
\small\texttt{19383273@brookes.ac.uk}\quad
\texttt{arast@brookes.ac.uk}\quad
\texttt{abradley@brookes.ac.uk}}
\date{}
\hypersetup{pdftitle={LightSEAD: Lightweight SE(3)-Aware Keypoint Detector through Selective Knowledge Distillation},pdfauthor={Mayowa Adebambo; Alexander D. Rast; Andrew Bradley},pdfsubject={Computer vision; keypoint detection; geometric supervision}}
\begin{document}
\maketitle
\begin{abstract}
The pursuit of improved performance in deep learning has often been accompanied
by increasing architectural complexity and computational demands. We explore an
alternative direction: achieving \textbf{more with less} through structured
supervision. We introduce LightSEAD, a lightweight SE(3)-aware keypoint detector
initialised with calibrated Harris pseudo-labels and enriched through
viewpoint-aware and appearance-aware supervision and selective knowledge
distillation. LightSEAD comprises only six convolutional layers and 84,289
learnable parameters, achieving competitive repeatability across HPatches, KITTI
06 and 4Seasons while demonstrating its downstream applicability to relative
pose estimation on MegaDepth-1500. Further optimised for mixed-precision
inference, LightSEAD Coarse achieves approximately 909 FPS on an NVIDIA RTX 4090
and 139 FPS on a Jetson Orin Nano Super.
\end{abstract}

\section{Introduction}
\label{sec:intro}

Real-time visual perception requires accurate features within limited compute
budgets. Repeatable geometric keypoints support image matching, camera calibration
and robotic applications such as simultaneous localisation and mapping (SLAM).
Handcrafted detectors, including Harris~\cite{harris1988combined},
FAST~\cite{rosten2006machine} and ORB~\cite{rublee2011orb}, use explicit
gradient or intensity criteria with low computational requirements, but their
repeatability can deteriorate under viewpoint and appearance variations.

Learned detectors such as DISK~\cite{tyszkiewicz2020disk} and
R2D2~\cite{revaud2019r2d2} optimise repeatable image locations through data-driven
supervision. XFeat~\cite{potje2024xfeat} demonstrates that learned local features
can also prioritise computational efficiency. These advances motivate using
richer supervision to improve geometry without increasing inference complexity.

\begin{figure}[t]
\centering
\includegraphics[width=0.92\columnwidth]{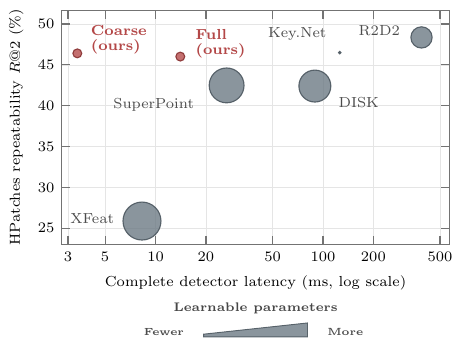}
\caption{Runtime--repeatability--parameter trade-off on HPatches. The axes show
complete-detector latency and $R@2$ under a top-1,000 detection ceiling;
bubble area is proportional to learnable parameter count. LightSEAD,
optimised for mixed-precision inference,
achieves a favourable trade-off between repeatability, computational efficiency
and architectural compactness.}
\label{fig:intro_parameter_efficiency}
\end{figure}

This motivates our central question: \textit{can structured geometric supervision
enable a compact learned keypoint detector to achieve competitive repeatability
and low inference latency without increasing architectural complexity?}
Figure~\ref{fig:intro_parameter_efficiency} shows why architectural
compactness, runtime and repeatability must be considered separately: Key.Net~\cite{barrosolaguna2019keynet}
has only 5,873 learnable parameters yet substantially higher measured latency,
whereas mixed-precision LightSEAD balances compactness, repeatability and
latency. We propose LightSEAD Full and LightSEAD Coarse as two inference
configurations of the same compact student trained with structured supervision.

\begingroup
\setlength{\parskip}{0pt}
The main contributions of this work are:
\begin{itemize}[topsep=0pt,partopsep=0pt,parsep=0pt,itemsep=0pt]
 \item\textbf{Competitive geometry with compact inference.} A six-convolution
detector preserves network-input resolution using dilated convolutions.
Its 84,289 parameters and 168.6\,kB FP16 weights support competitive repeatability.

\item\textbf{Calibration-guided pseudo-label supervision.} Known synthetic
corners calibrate Harris parameters to provide structured pseudo-labels,
initialising K from random weights over 20 epochs.

\item\textbf{Geometrically informed knowledge distillation.} We present a pair-based teacher
that uses viewpoint, appearance, and geometric verification signals to supervise
a single-image student. At inference, the student requires no paired frames,
depth, pose, or temporal history. We evaluate the training signals through the K
$\rightarrow$ KV $\rightarrow$ KVA progression.

\item\textbf{Efficient cross-domain and embedded detection.} We evaluate
LightSEAD on KITTI 06, 4Seasons and HPatches, testing transfer from driving-oriented
training to indoor scenes and less constrained viewpoints. Mixed-precision
LightSEAD Coarse achieves 7.206\,ms per
image (about 139\,FPS) on KITTI 06 with a Jetson Orin Nano Super.
\end{itemize}
\endgroup

\section{Related Work}
\label{sec:RW}
\subsection{Handcrafted Keypoint Detectors}
Classical detectors encode explicit notions of local saliency. Harris and
Stephens score the image-gradient second-moment matrix~\cite{harris1988combined},
whereas Shi--Tomasi selects locations with a large minimum eigenvalue to support
tracking~\cite{shi1994good}. FAST instead applies learned intensity tests on a
pixel ring for rapid corner decisions~\cite{rosten2006machine}; ORB augments
FAST-based detection with orientation and a rotated binary
descriptor~\cite{rublee2011orb}. Their efficiency and predictable computation
remain valuable in visual odometry, SLAM, and resource-constrained robotics.
However, fixed response criteria offer limited adaptation to substantial
viewpoint or appearance changes. LightSEAD retains their useful inductive bias:
calibrated Harris detections provide its initial pseudo-labels, rather than its
deployed inference mechanism.

\subsection{Learned Keypoint Detectors}
Learned detectors differ chiefly in how they define useful and repeatable
points. TILDE learns stability across temporal illumination and weather
variation~\cite{verdie2015tilde}, while SuperPoint bootstraps joint detection
and description from synthetic corners and homographic
adaptation~\cite{detone2018superpoint}. Key.Net combines handcrafted gradient
filters with shallow learned filters and multi-scale repeatability
objectives~\cite{barrosolaguna2019keynet}. Other methods couple detection more
directly to matching utility: R2D2 predicts repeatability and
reliability~\cite{revaud2019r2d2}, DISK optimises local features using policy
gradients~\cite{tyszkiewicz2020disk}, and SiLK learns detection and description
through a deliberately simple differentiable formulation~\cite{gleize2023silk}.
DeDoDe instead decouples detection from description and supervises detections
using 3D-consistent tracks~\cite{edstedt2024dedode}, an especially relevant
precedent for detector-focused geometric supervision. This distinction motivates
transferring geometric knowledge without retaining paired search at inference.

Efficiency-oriented designs show that learned local features need not require
heavy inference. ALIKE introduces differentiable sub-pixel detection in a
lightweight detector--descriptor network~\cite{zhao2023alike}; ALIKED uses a
sparse deformable descriptor head to reduce descriptor
computation~\cite{zhao2023aliked}. XFeat revisits backbone resolution, channel
allocation, and sparse/semi-dense extraction for resource-efficient
matching~\cite{potje2024xfeat}. Key.Net has fewer learnable parameters than
LightSEAD, highlighting that parameter count and measured latency are distinct.
LightSEAD explores a different point in this
design space: a detector-only, 84,289-parameter student initialised with
calibrated handcrafted pseudo-labels, then trained with viewpoint- and
appearance-aware representations and geometrically verified selective
distillation, retaining neither auxiliary representations nor paired search at inference.

\begingroup
\graphicspath{{sad_cvpr_methodology/}}
\section{Methodology}
\label{sec:sad_method}

\subsection{Framework Overview}
LightSEAD combines calibrated pseudo-label initialisation, geometric and
appearance representation learning, and selective paired distillation.
An initial keypoint detector $K$ supplies source queries and independent target
detections. Viewpoint and appearance branches $V$ and $A$ provide separate
32-dimensional representations for proposing additional locations. A frozen
teacher combines these proposals; reference geometry selects detected endpoints
for student supervision. The deployed student retains only an encoder and a
scalar detection head. Temporal geometry is used during training, while
inference operates on a single image without depth, pose, or previous frames.

The training framework draws images from KITTI~\cite{geiger2012kitti} sequences
00--05, 07, 09 and 10; 4Seasons~\cite{wenzel2020fourseasons} Neighbourhood 1,
Neighbourhood 3 and Business Campus; Mapillary Vistas~\cite{mapillaryimages}; Oxford
RobotCar~\cite{maddern2016robotcar}. Synthetic
homographies generate geometric training pairs from source images rather than
constitute an independent image dataset. Stages use different subsets: K uses
all image sources, while V and student distillation use KITTI 00--05/07 and
Neighbourhood geometry, with Mapillary also supplying homography sources. A uses
photometric pairs from the training image sources. KITTI 06 and May Parking
Garage are evaluation and validation sequences, excluded from optimisation.

Let $I_s,I_t$ denote source and target images and $p_s$ a source point.
We distinguish its geometric projection $\hat p_t$ from an actual detected target
coordinate $q_t$. Coordinates used for geometric verification are native-image
pixels; heatmaps are constructed on the resized network grid.

Figure~\ref{fig:sad_architecture} separates the frozen, pair-based teacher used
to construct supervision from the single-image network retained at deployment.

\begin{figure*}[!t]
\centering
\includegraphics[width=\textwidth]{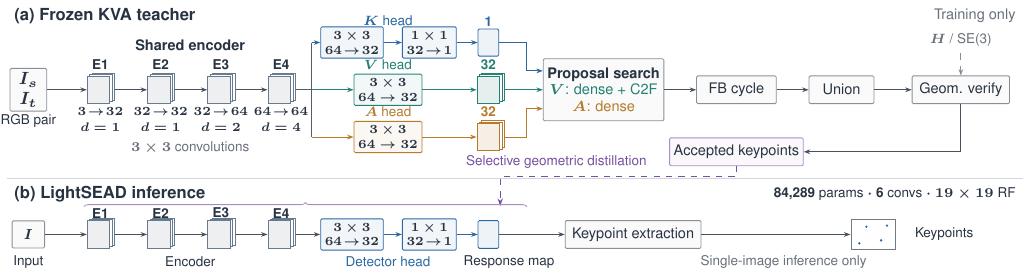}
\caption{Frozen KVA teacher and deployed LightSEAD. (a) A shared encoder provides
$K$ detections and 32-D $V/A$ features. (b) Selective distillation yields
a single-image detector with six convolutions and 84,289 parameters.}
\label{fig:sad_architecture}
\end{figure*}

\subsection{Calibration-Guided Pseudo-Label Initialisation}
Calibration uses 120 outlined geometric images, with 20 examples
each of triangles, squares, rectangles, pentagons, hexagons, and circles.
Polygon vertices provide known corners for systematic calibration; corner-free
circles expose spurious responses. The data comprise 80 calibration and 40 validation
images. We evaluate 195 classical detector configurations, including
90 Harris configurations, and select parameters using calibration data only.
The candidates use Harris~\cite{harris1988combined}, FAST~\cite{rosten2006machine},
ORB~\cite{rublee2011orb} and Shi--Tomasi~\cite{shi1994good} corner criteria.
Predictions and reference corners are greedily paired in ascending distance
within three pixels without endpoint reuse. With pooled $F_1$, mean matched
localisation error $e$, and false positives per image $f$, the criterion is
\begin{equation}
 L_{\mathrm{cal}}=1-F_1+0.25\frac{e}{3}+0.25\frac{f}{10}.
\end{equation}
The criterion balances precision/recall, localisation precision and excess
detections; the weights specify this trade-off, without an optimality claim.
No matches incur $e=3$. The selected Harris configuration uses a five-pixel
block, derivative aperture five, $k=0.04$, and a response threshold of $0.05$
times the image maximum, with $3\times3$ local-maxima extraction. These settings
are fixed before generating native-image pseudo-labels for $K$.

\begin{table}[t]
\centering
\footnotesize
\setlength{\tabcolsep}{2pt}
\renewcommand{\arraystretch}{1.1}
\begin{tabular}{@{}lrrrr@{}}
\toprule
Detector & $F_1\uparrow$ & \shortstack{Loc. err.\\(px)$\downarrow$} & \shortstack{FP/img.\\$\downarrow$} & \shortstack{Circle\\FP/img.$\downarrow$} \\
\midrule
Harris (default) & 0.050 & 1.481 & 137.700 & 113.167 \\
Harris (cal.) & \textbf{0.646} & 1.169 & \textbf{3.850} & \textbf{7.000} \\
FAST (default) & 0.291 & 1.800 & 8.925 & 7.333 \\
FAST (cal.) & 0.291 & 1.800 & 8.925 & 7.333 \\
ORB (default) & 0.051 & \textbf{0.866} & 129.475 & 120.167 \\
ORB (cal.) & 0.103 & 0.876 & 60.575 & 41.000 \\
Shi--Tomasi (default) & 0.100 & 1.792 & 65.300 & 66.333 \\
Shi--Tomasi (cal.) & 0.191 & 1.649 & 18.050 & 19.333 \\
\bottomrule
\end{tabular}
\caption{Held-out synthetic validation on 40 images after calibration on a
separate 80-image set. Circle FP/img. averages detections across six circles
with zero annotated corners; bold denotes the best value per column.}
\label{tab:synthetic_calibration}
\end{table}

Table~\ref{tab:synthetic_calibration} reports the held-out results. Harris
was selected with the lowest calibration objective ($0.528$), before validation.
It achieves the highest validation $F_1$ and the fewest overall and circle
false positives among these configurations; default ORB has the lowest matched
localisation error. Circle false positives are not a separate selection term;
they contribute to the aggregate false-positive count in the objective.

\subsection{Keypoint Head Supervision}
The encoder (Fig.~\ref{fig:sad_architecture}) contains four stride-one $3\times3$ convolutions with channels
$3\rightarrow32\rightarrow32\rightarrow64\rightarrow64$ and dilations
$1,1,2,4$. Each is followed by eight-group normalisation~\cite{wu2018group} and ReLU. A head applies
a $3\times3$, $64\rightarrow32$ convolution, group normalisation and ReLU, then a
$1\times1$, $32\rightarrow1$ convolution. The output $z_K(I)$ contains logits.
Input RGB is scaled to $[0,1]$ and downscaled, without upscaling, to a maximum
long side of 960. Native labels are mapped by the actual rounded resize ratios.

For centers $\mathcal Q$, the target is the maximum of Gaussians of standard
deviation $1.5$ network pixels, each retained on its clipped finite floor/ceil
radius-four rectangular footprint. Let $Y$ be this soft target,
$\mathcal P=\{u:Y(u)\ge0.05\}$, and $\mathcal N$ its complement. Using
binary cross-entropy with logits $\ell$, the original $K$ objective is
\begin{equation}
\begin{split}
 L_K={}&\frac{1}{2|\mathcal P|}\sum_{u\in\mathcal P}\ell(z_K(u),Y(u))\\
       &+\frac{1}{2|\mathcal N|}\sum_{u\in\mathcal N}\ell(z_K(u),Y(u)).
\end{split}
\end{equation}
If one region is empty, its term is omitted and the remaining mean receives
unit weight. Initial training uses 5,000 training and 1,500 validation images
from the real-image sources above, separately from synthetic calibration.
AdamW~\cite{loshchilov2019adamw} uses
learning rate $10^{-3}$, zero weight decay, batch size one and 20 epochs. The
parameters with minimum validation loss initialise subsequent stages.

\subsection{Viewpoint-Aware Geometric Supervision}
With the encoder and $K$ frozen, $V$ learns a separate $64\rightarrow32$
convolution--normalisation--ReLU representation. Its scalar output layer is
frozen and does not define descriptor supervision. We sample 64 corresponding
locations per pair, bilinearly sample their descriptors, and normalise each
descriptor. For normalised source and target vectors $a_i,b_i$, define
$S_{ij}=a_i^\top b_j/\tau$, with $\tau=0.1$. The representation objective is
\begin{equation}
\begin{split}
 L_{\mathrm{rep}}=-\frac{1}{2N}\sum_{i=1}^{N}
 \bigg(&\log\frac{e^{S_{ii}}}{\sum_j e^{S_{ij}}}\\
       &+\log\frac{e^{S_{ii}}}{\sum_j e^{S_{ji}}}\bigg),\quad N=64.
\end{split}
\end{equation}
All off-diagonal candidates act as negatives. There are no cross-batch
negatives or explicit pose-regression targets.

Homography pairs warp both the image and source coordinates by a known
source-to-target homography, with validity derived from a warped support mask.
Real pairs use stereo camera-Z depth and camera-to-world poses. For source
intrinsics $K_s$ and depth $d_s$, we compute
\begin{align}
 X_s&=d_s(p_s)K_s^{-1}\widetilde p_s,\\
 T_{t\leftarrow s}&=T_{w\leftarrow t}^{-1}T_{w\leftarrow s},\\
 X_t&=R_{t\leftarrow s}X_s+t_{t\leftarrow s},\qquad
 \hat p_t=\pi(K_tX_t).
\end{align}
We reject invalid source depth, nonpositive target depth, and out-of-bounds
projections. Target depth is estimated independently. A valid, known target
depth is required; points with
$Z_t>d_t(\hat p_t)+\max(1\,\mathrm m,0.05d_t(\hat p_t))$ are rejected as likely
occluded. This is a one-sided non-occlusion check, not a symmetric depth-error
bound. Stereo uses native-resolution reverse-consistent SGBM~\cite{hirschmuller2008stereo} without filling.
KITTI and validated 4Seasons poses use camera-to-world conventions; 4Seasons
poses are in a local VIO world with nominal metric scale. They and stereo depth
are reference estimates rather than exact depth/pose ground truth.

Motion-stratified SE(3) supervision uniformly samples small, moderate and large
translation or turning ranks (Table~\ref{tab:main_bins}), encouraging $V$'s
geometric consistency across viewpoint changes. Real scenes complement synthetic
homographies with depth-dependent variation. Groups contain one anchor and three
independently checked targets. Section~\ref{sec:motion_results} evaluates
repeatability across these motion magnitudes.
\begin{table}[H]
\centering
\caption{Anchor-relative motion bins and group validity criteria.}
\label{tab:main_bins}
\begingroup\footnotesize
\setlength{\tabcolsep}{5pt}
\renewcommand{\arraystretch}{1.0}
\begin{tabular}{@{}lccc@{}}
\toprule
Motion & Rank 1 & Rank 2 & Rank 3 \\
\midrule
Straight & $[1,3)$ m & $[3,6)$ m & $[6,10]$ m \\
Turn & \mbox{$[2,5)^\circ$} & \mbox{$[5,10)^\circ$} & \mbox{$[10,20]^\circ$} \\
\bottomrule
\end{tabular}
\endgroup
\par\smallskip
{\footnotesize\raggedright
Straight groups require $|\psi|<2^\circ$ throughout. Turns use same-sign yaw
and chronological targets. KITTI/Garage limits are $10/12$\,m translation
and $30/90$ frames between targets. Continuous groups require at least
64 correspondences independently at each rank.\par}
\end{table}

Training samples homographies and real geometry equally and balances real datasets
and motion types. The selected $V$ parameters are
from epoch 19 of 20, with 2,000 Adam~\cite{kingma2015adam} updates per epoch and learning rate
$10^{-3}$.

\subsection{Appearance Consistency Supervision}
The teacher's $A$ branch uses its own 32-dimensional
convolution--normalisation--ReLU representation and the same symmetric
classification objective. The encoder, $K$, $V$, and the scalar $A$ output
remain frozen. $A$ uses photometric variants at unchanged coordinates,
including exposure, gamma, white balance, contrast, saturation, darkening,
noise, haze, and mild blur. Positives require locally observable target
structure, but not rediscovery by target $K$. Raw-image groups remain disjoint
between training and calibration. Training and calibration contain 9,927 and
1,122 photometric pairs, respectively. Training uses 40,000 Adam updates at
$10^{-3}$ with a mild-to-hard curriculum; the teacher uses the epoch-14
parameters.

\subsection{Geometric Teacher Construction}
For each source $K$ query, dense $V$ and dense $A$ independently search target
feature grids by cosine similarity and perform backward search. A contextual
$V$ stream pools normalised features by two, selects eight coarse peaks, and
searches $21\times21$ fine windows using original fine descriptors. This is
frozen search added after representation training. Each stream requires
forward--backward error at most five native pixels.

The teacher preserves every target $K$ point, appends $V$ proposals farther
than five pixels from $K$, removes exact cross-stream duplicates, and finally
appends $A$ proposals farther than five pixels from the $K/V$ union. Geometry
does not enter these operational searches. Subsequently, for a valid source
teacher detection $p$, the nearest actual target detection $q$ is accepted if
\begin{equation}
 \|q-g_{t\leftarrow s}(p)\|_2\le5\ \text{native pixels}.
\end{equation}
Forward and reverse verification are independent; their accepted index pairs are
unioned and deduplicated. One-way and many-to-one pairs are retained. Crucially,
supervision centers on the actual verified detection $q$, rather than replacing
it with its geometric projection. Failed candidates remain unsupervised. The
five-pixel tolerance accommodates localisation uncertainty and imperfect
depth/pose references.

\subsection{Selective Knowledge Distillation}
The student initialises from the trained $K$ checkpoint and fine-tunes its
encoder and detector head on 16,000 scheduled image pairs:
8,006 homography and 7,994 real-SE(3) pairs, from the $V$ training sources.
The frozen KVA teacher is applied in both directions before geometric verification.
Known homographies, or stereo depth and camera poses for real SE(3) pairs,
verify correspondences; proposals alone are not ground truth. Three objectives
use geometrically accepted detections.

\noindent\textbf{Anchor loss ($L_{\mathrm{anchor}}$).}
Project an original $K$ point into the opposite view and supervise a verified
target detection within five pixels, preferring valid target-$K$ detections
over verified KVA rescue detections.

\noindent\textbf{Pair loss ($L_{\mathrm{pair}}$).}
Supervise both detected endpoints of every geometrically verified KVA
correspondence, including valid $V$/$A$ detections.

\noindent\textbf{Preservation loss ($L_{\mathrm{preserve}}$).}
Reinforce verified original-$K$ correspondences on real SE(3) pairs, with no
support on homography pairs.

For term $c$, pair $b$, and view $v$, let $Y_{bvc}$ be its Gaussian-MAX target
centred on selected detections and $M_{bvc}$ the union of finite target footprints intersected with valid
image support. Pixels outside this support are ignored, not negative labels.
For a logical batch of four pairs,
\begin{equation}
 L_c=\frac12\sum_{v=0}^{1}
 \frac{\sum_{b=1}^{4}\sum_u M_{bvc}(u)
 \ell(z_\theta(I_{bv})(u),Y_{bvc}(u))}
 {\max(1,\sum_{b=1}^{4}\sum_u M_{bvc}(u))}.
\end{equation}
The combined objective is
\begin{equation}
 L_{\mathrm{LightSEAD}}=L_{\mathrm{anchor}}+L_{\mathrm{pair}}
                  +0.25L_{\mathrm{preserve}}.
\end{equation}
This couples label construction across views. AdamW has zero weight decay;
encoder-block learning rates are $(2.5,5,7.5,10)\times10^{-6}$ and head learning
rate is $10^{-4}$. Four pairs produce eight sequential forwards and one
accumulated update. Optimisation runs for 4,000 accumulated updates. Inference
uses only the single-image student, without KVA or geometric verification.

\subsection{LightSEAD Inference}
The six-convolution, 84,289-parameter network produces a dense confidence map
in one forward pass. Its nominal receptive field is $19\times19$; group normalisation
adds whole-input dependence. Sigmoid thresholding at 0.50
extracts every above-threshold pixel, retaining neighbouring high-confidence
responses; coordinates are then rescaled, with no point cap. NMS remains
optional post-processing for sparse extraction according
to downstream needs; it does not alter the learned architecture.

LightSEAD Coarse uses the same trained weights with network input subsampled
to half the dimension along each image axis. It maps pixel centres back using
$(x_c+0.5)W/W_c-0.5$ before native-coordinate
rescaling. As a deployment optimisation, TensorRT preserves the network and
its normalisation using explicit mixed FP16 with FP32 normalisation and I/O.

\endgroup

\begingroup
\graphicspath{{LightSEAD_Experiments_Results_Compact/}}
\makeatletter
\def\input@path{{LightSEAD_Experiments_Results_Compact/}}
\makeatother
\section{Experiments and Results}
\subsection{Evaluation Protocol}
We evaluate detection repeatability and cost on HPatches~\cite{balntas2017hpatches}
(580 pairs: 285 illumination, 295 viewpoint), KITTI06~\cite{geiger2012kitti}
and 4Seasons May Parking Garage~\cite{wenzel2020fourseasons}
(\texttt{2021-05-10\_19-15-19}; 512 ranked pairs each).
KITTI06/Garage assess driving-domain performance; HPatches assesses
cross-domain transfer.

Each detector supplies points in both views. Homography (HPatches) or calibrated
pose/stereo depth (driving data) projects evaluable source points to the target.
We pool both directions as $R_\tau=100\sum h_\tau/\sum|V|$ for $\tau=2,5$
native pixels; $D/E$ average directional appearances, not unique images.
A repeatability hit occurs when an evaluable source detection projects within
$\tau$ pixels of any target detection, without enforcing one-to-one correspondence.
Full and Coarse share the same weights and use threshold-based extraction at
0.50; Coarse halves the network input resolution. Original K uses threshold
0.65 and radius-one spatial suppression. Native evaluation retains each
method's detections; budgeted evaluation ranks scores under top-1,000/1,500
ceilings without padding. Cap-specific latency is not measured. KV/KVA teacher
coverage instead uses fixed K source queries. Motion ranks comprise independently
evaluable cohorts rather than common tracks.

\noindent\textbf{Latency protocol.}\enspace Batch-one totals span preprocessed
CPU RGB to CPU coordinates/scores, including conversion, transfer, inference and
extraction, but excluding shared loading, decode/resize, matching and geometry.
RTX 4070 Laptop supports comparative benchmarking, RTX 4090 assesses
high-throughput/backend performance, and Jetson Orin Nano Super evaluates
embedded deployment. Learned baselines use PyTorch FP32;
ORB~\cite{rublee2011orb} and Harris~\cite{harris1988combined} use CPU. LightSEAD
also uses TensorRT mixed-precision inference with FP16 weights and accelerated
kernels, retaining selected FP32 operations. Both configurations separate
architectural from deployment gains; competing learned detectors use PyTorch
FP32. SuperPoint~\cite{detone2018superpoint}, R2D2~\cite{revaud2019r2d2},
DISK~\cite{tyszkiewicz2020disk} and ORB use detector-only paths.
The XFeat timing excludes descriptor post-processing but includes
descriptor-feature extraction required for its native reliability ranking.
Key.Net uses the same
complete-detector boundary. 

\subsection{Repeatability}
\subsubsection{Native-count and Point-budget Repeatability}
\begin{table*}[t]
\centering
\fontsize{6.8}{7.7}\selectfont
\setlength{\tabcolsep}{2.7pt}
\renewcommand{\arraystretch}{1.03}
\begin{tabular}{lrrrrrrrrr}
\toprule
& \multicolumn{3}{c}{HPatches} & \multicolumn{3}{c}{KITTI06} & \multicolumn{3}{c}{May Garage} \\
\cmidrule(lr){2-4}\cmidrule(lr){5-7}\cmidrule(lr){8-10}
Method & Native & @1k & @1.5k & Native & @1k & @1.5k & Native & @1k & @1.5k \\
\midrule
K          & 42.43/65.85 & 39.88/60.91 & 41.00/62.95 & 62.10/87.16 & 62.10/87.16 & 62.10/87.16 & 27.54/82.78 & 27.54/\textbf{82.78} & 27.54/82.78 \\
Full (ours) & \textbf{60.30}/75.39 & 46.02/59.18 & 48.75/62.72 & \textbf{79.27}/\textbf{91.92} & \textbf{72.88}/88.44 & \textbf{75.85}/\textbf{90.24} & \textbf{52.42}/\textbf{88.98} & \textbf{30.86}/80.28 & \textbf{37.03}/\textbf{83.60} \\
Coarse (ours) & 53.40/70.47 & 46.41/62.68 & 48.31/65.04 & 71.83/88.48 & 70.57/87.85 & 71.83/88.48 & 34.92/72.11 & 30.34/67.80 & 32.82/70.47 \\
\midrule
SuperPoint & 40.42/66.40 & 42.48/62.48 & 41.87/64.93 & 49.89/77.20 & 51.67/76.96 & 49.95/77.20 & 13.47/58.59 & 13.47/58.59 & 13.47/58.59 \\
ORB        & 45.80/62.10 & 45.80/62.10 & 45.80/62.10 & 69.57/88.64 & 69.57/\textbf{88.64} & 69.57/88.64 & 29.43/74.87 & 29.43/74.87 & 29.43/74.87 \\
Harris     & 42.96/65.97 & 41.64/63.20 & 42.21/64.40 & 62.88/87.49 & 62.88/87.49 & 62.88/87.49 & 28.17/82.06 & 28.17/82.06 & 28.17/82.06 \\
Key.Net    & 47.65/70.14 & 46.50/\textbf{67.20} & 48.67/\textbf{68.76} & 53.94/77.48 & 61.80/83.26 & 60.35/82.38 & 18.44/61.30 & 18.20/61.23 & 18.44/61.30 \\
R2D2       & 51.11/76.09 & \textbf{48.36}/65.77 & \textbf{49.30}/68.05 & 53.63/82.39 & 58.58/78.97 & 57.77/79.62 & 22.32/64.83 & 16.75/48.52 & 18.09/52.35 \\
DISK       & 47.03/\textbf{79.95} & 42.40/58.44 & 44.15/63.59 & 55.22/90.67 & 53.12/75.65 & 54.52/80.14 & 25.97/85.20 & 20.74/78.00 & 23.40/82.47 \\
XFeat      & 35.68/68.78 & 25.87/43.86 & 28.54/50.52 & 42.03/85.11 & 30.98/58.39 & 34.36/66.78 & 22.66/80.92 & 17.35/70.38 & 20.21/76.92 \\
\bottomrule
\end{tabular}
\caption{Native-count and point-budgeted repeatability. Values are $R_2/R_5$ (\%) on HPatches, KITTI06, and May Garage. Native evaluation uses each detector's own detections; @1k and @1.5k impose top-ranked detection ceilings without padding. Bold denotes the best value within each dataset and point-count protocol.}
\label{tab:main_native}
\label{tab:main_caps}
\end{table*}

Despite using only six convolutions and 84,289 parameters, LightSEAD Full
achieves the highest native $R_2$ across all three datasets: 60.30\% on
HPatches, 79.27\% on KITTI06 and 52.42\% on May Garage
(Table~\ref{tab:main_native}). This performance supports the central motivation:
structured supervision can develop competitive geometric features within a
compact backbone. Coarse offers a complementary operating point, running
3.9--4.3$\times$ faster in FP32 while preserving substantial repeatability
(Fig.~\ref{fig:main_tradeoff}).
\par\noindent\textbf{Point-budgeted repeatability.}\enspace
Fixed ceilings complement native evaluation by testing score-ranked selection.
At top-1,000, Full maintains leading $R_2$ on both driving datasets, while
Full and Coarse remain competitive with R2D2 and Key.Net on HPatches.
Coarse exceeds Full's HPatches $R_5$ by 3.50/2.32 percentage points at
top-1,000/1,500 (Table~\ref{tab:main_caps}), indicating greater robustness to
score truncation. R2D2 leads budgeted HPatches $R_2$
(48.36/49.30), whereas Key.Net leads $R_5$ (67.20/68.76).

\begin{figure}[H]
\centering
\includegraphics[width=0.80\columnwidth]{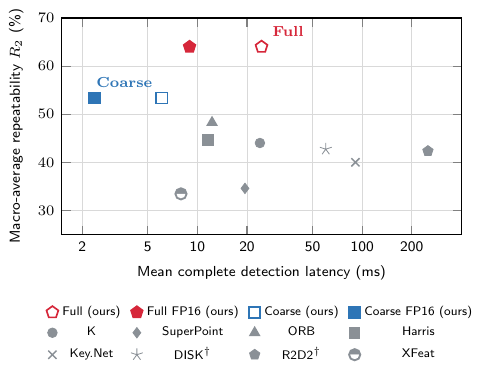}
\caption{Macro-average $R_2$--latency trade-off on an RTX 4070 Laptop GPU
across HPatches, KITTI06, and May Garage; Tables~\ref{tab:main_native} and
\ref{tab:main_native_runtime} give dataset-level results. Full/Coarse denote
LightSEAD; FP16 denotes mixed precision, other configurations use FP32.}
\label{fig:main_tradeoff}
\end{figure}

Native and capped results capture complementary density and ranking effects:
Full retains higher driving $R_2$, while Coarse offers efficient capped detection.

\subsubsection{Cross-Domain Generalisation on HPatches}
LightSEAD transfers from driving-oriented training to HPatches' indoor and
outdoor scenes. Full leads independent-query
$R_2$ under both illumination and viewpoint changes, as well as viewpoint
$R_5$ (Fig.~\ref{fig:main_hpatches_comparison}). DISK leads illumination
$R_5$, demonstrating tolerance-dependent strengths across methods.

Coarse maintains competitive half-resolution recall, while Full's larger
advantage under viewpoint change highlights the value of spatial detail.
The fixed-query KV/KVA rows show how richer teacher proposals extend target
coverage. Figure~\ref{fig:main_hpatches_indoor} illustrates transfer to an indoor
illumination sequence.
\par
\begin{figure}[t]
\centering
\includegraphics[width=\linewidth]{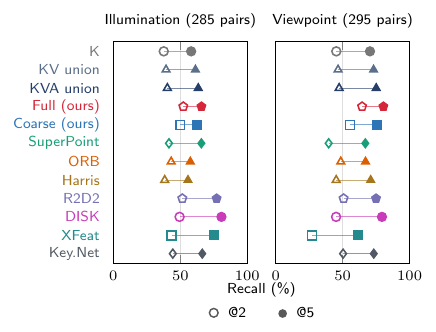}
\caption{HPatches native own-source recall for ten detectors under
illumination and viewpoint changes.
Hollow/filled markers denote @2/@5; each row identifies one detector.
Full/Coarse (ours) denote LightSEAD. KV/KVA union rows use fixed K queries;
Key.Net uses a 4,096-point ceiling.}
\label{fig:main_hpatches_comparison}
\end{figure}
\par
\begin{figure}[tbp]
\centering
\includegraphics[width=\linewidth]{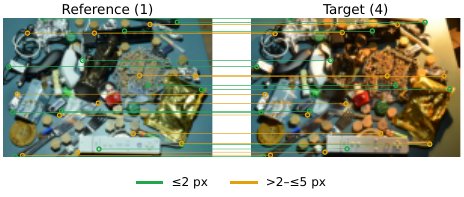}
\caption{Indoor HPatches \texttt{i\_objects}, frames 1$\to$4:
Full LightSEAD under illumination change. The supplied homography associates
twelve links at $\le2$ px and twelve at $>2$--$\le5$ px.}
\label{fig:main_hpatches_indoor}
\end{figure}

\subsubsection{Translation and Turning Motion}
\label{sec:motion_results}
\begin{figure}[t]
\centering
\includegraphics[width=0.92\linewidth]{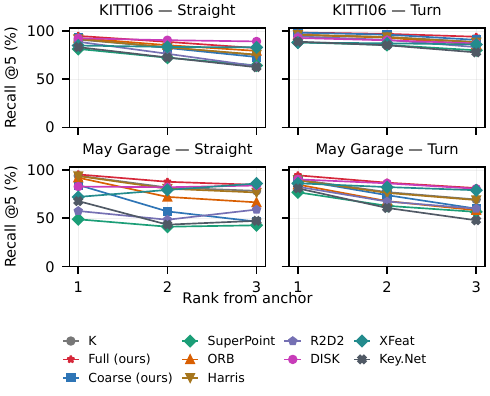}
\caption{Native $R_5$ on identical rank cohorts; colours and shapes identify
methods. Full/Coarse (ours) denote LightSEAD.}
\label{fig:main_ranked_comparison}
\end{figure}
The ranked experiments assess increasing camera motion
(Fig.~\ref{fig:main_ranked_comparison}). Full retains stronger
$R_5$ than Coarse, with a widening gap across ranks in all four groups,
especially on Garage. Native spatial detail helps preserve localisation evidence
under larger geometric changes, complementing viewpoint-aware training.
DISK leads KITTI straight ranks 2/3 and XFeat leads Garage straight rank 3,
showing that strengths also depend on motion and tolerance.
Figure~\ref{fig:main_pairs} illustrates the corresponding geometric settings.
\begin{table*}[t]
\centering
\fontsize{7.3}{8.1}\selectfont
\setlength{\tabcolsep}{3.0pt}
\renewcommand{\arraystretch}{1.02}
\begin{tabular}{@{}lrrrrrrrrrrr@{}}
\toprule
& Params & Weights & \multicolumn{3}{c}{RTX 4070 Laptop} & \multicolumn{3}{c}{RTX 4090} & \multicolumn{3}{c}{Jetson Orin Nano Super} \\
\cmidrule(lr){4-6}\cmidrule(lr){7-9}\cmidrule(lr){10-12}
Method/backend & (M) & (kB) & H & K & G & H & K & G & H & K & G \\
\midrule
LightSEAD Full PT32  & 0.084 & 337.2 & 36.805 & 16.935 & 19.758 & 17.490 & 7.711 & 9.256 & 146.284 & 66.972 & 77.489 \\
LightSEAD Full TRT16 & 0.084 & 168.6 & 14.059 & 5.599 & 7.225 & 3.950 & 1.860 & 2.430 & 47.110 & 22.371 & 24.540 \\
LightSEAD Coarse PT32 & 0.084 & 337.2 & 9.553 & 4.053 & 4.619 & 4.290 & 1.980 & 2.190 & 38.341 & 18.147 & 21.156 \\
LightSEAD Coarse TRT16 & 0.084 & 168.6 & \textbf{3.411} & \textbf{1.809} & \textbf{1.896} & \textbf{1.640} & \textbf{1.100} & \textbf{1.180} & \textbf{13.753} & \textbf{7.206} & \textbf{7.761} \\
\midrule
SuperPoint         & 1.301 & 5203.5 & 26.571 & 18.785 & 12.977 & 9.560 & 3.620 & 4.360 & 106.064 & 46.787 & 53.584 \\
ORB                & -- & -- & 14.972 & 12.776 & 9.060 & 9.620 & 5.490 & 5.800 & 27.622 & 13.898 & 14.398 \\
Harris             & -- & -- & 15.434 & 11.416 & 7.912 & 9.850 & 4.400 & 5.080 & 25.913 & 13.426 & 14.544 \\
Key.Net            & 0.006 & 23.5 & 125.890 & 70.380 & 76.770 & 68.160 & 47.000 & 48.550 & 663.784 & 323.573 & 351.236 \\
DISK detector$^\dagger$ & 1.092 & 4369.5 & 89.432 & 43.539 & 47.390 & 18.650 & 9.200 & 9.990 & 620.069 & 293.282 & 320.763 \\
R2D2 detector$^\dagger$ & 0.484 & 1937.5 & 387.601 & 171.434 & 193.023 & 100.190 & 45.010 & 49.710 & 2274.336 & 1043.619 & 1158.989 \\
XFeat              & 1.545 & 6179.0 & 8.300 & 7.910 & 7.642 & 7.460 & 7.000 & 7.060 & 30.160 & 23.698 & 24.067 \\
\bottomrule
\end{tabular}
\caption{Parameter footprint and complete latency (ms) on HPatches (H),
KITTI06 (K), and May Garage (G). PT32/TRT16 denote PyTorch FP32/TensorRT FP16;
external learned methods use PyTorch FP32, while ORB/Harris run on CPU.
$^\dagger$Detector-only timing; dashes denote unmeasured configurations.}
\label{tab:main_native_runtime}
\end{table*}

\begin{figure}[t]
\centering
\includegraphics[width=\linewidth,trim=0 0 30bp 0,clip]{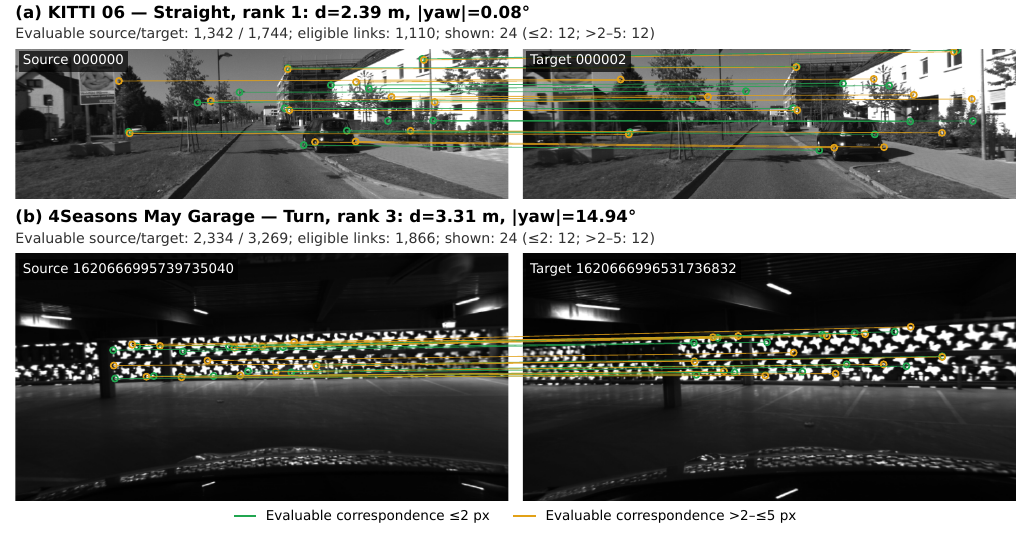}
\caption{Evaluable Full examples for KITTI straight rank1 and Garage turn
rank3, each showing twelve reference-assisted links $\le2$ px and twelve
$>2$--$\le5$ px.}
\label{fig:main_pairs}
\end{figure}
\subsubsection{Student Distillation Ablation}
\label{sec:student_distillation_ablation}
Distillation tests richer supervision within the same compact architecture
(Table~\ref{tab:student_distillation}).
Starting from calibrated Harris pseudo-labels, geometrically verified K-only
distillation raises $R_2$ by 11.05--22.96 percentage points across datasets.
Viewpoint-aware KV targets add 1.90--6.08 points at $R_2$ and 1.25--4.07 at
$R_5$ through additional geometric proposals.
Appearance-aware KVA adds a smaller, complementary refinement of at most
0.29 points. Every stage retains the six-convolution, 84,289-parameter student:
the additional teacher representations and search enrich training rather than
inference. Optimisation budgets differ across stages, so this progression
supports the supervision strategy rather than strict causal isolation of each
component.
\begin{table}[H]
\centering
\footnotesize
\setlength{\tabcolsep}{2.8pt}
\renewcommand{\arraystretch}{1.05}
\begin{tabular}{lccc}
\toprule
Supervision & HPatches & KITTI 06 & May Garage \\
 & $R_2/R_5$ (\%) & $R_2/R_5$ (\%) & $R_2/R_5$ (\%) \\
\midrule
K baseline       & 42.43/65.85 & 62.10/87.16 & 27.54/82.78 \\
K-only distilled & 55.22/72.00 & 73.15/90.66 & 50.50/84.87 \\
KV distilled     & 60.10/75.10 & 79.23/91.91 & 52.40/88.94 \\
KVA distilled    & \textbf{60.30}/\textbf{75.39}
                 & \textbf{79.27}/\textbf{91.92}
                 & \textbf{52.42}/\textbf{88.98} \\
\bottomrule
\end{tabular}
\caption{Student-distillation ablation: native $R_2/R_5$ (\%). All students
share the same 84,289-parameter architecture; bold denotes the column best.}
\label{tab:student_distillation}
\end{table}

\subsection{Relative Pose Estimation}
As LightSEAD is a standalone detector, we pair it and Key.Net with the same
fixed pretrained DISK~\cite{tyszkiewicz2020disk} descriptor to further assess
downstream geometric utility on MegaDepth-1500~\cite{li2018megadepth}
(Table~\ref{tab:megadepth_pose}). DISK serves as an external descriptor for this
evaluation; other learned pipelines retain their native descriptors.

Full achieves the highest AUC at $5^\circ$ and $10^\circ$, reaching 45.87\%
and 58.27\%, and remains competitive at $20^\circ$, where R2D2 leads.
It also exceeds Key.Net at all three thresholds with the shared descriptor.
These results demonstrate downstream geometric utility for a detector trained
without an explicit pose-estimation objective. Coarse remains competitive with
an average of only 1,518 keypoints per image, substantially fewer than several
competing pipelines. Descriptor compatibility and retained counts contribute
to these pipeline-level outcomes.

\begin{table}[H]
\centering
\fontsize{7.4}{8.3}\selectfont
\setlength{\tabcolsep}{3.2pt}
\renewcommand{\arraystretch}{1.04}
\begin{tabular}{@{}lrrrr@{}}
\toprule
Pipeline & \shortstack{Points/\\image} & \mbox{AUC@$5^\circ$} & \mbox{AUC@$10^\circ$} & \mbox{AUC@$20^\circ$} \\
\midrule
LightSEAD Full + DISK   & 4,090 & \textbf{45.87} & \textbf{58.27} & 68.04 \\
Native DISK             & 4,096 & 38.75 & 55.06 & 68.17 \\
Native R2D2             & 4,094 & 36.05 & 53.61 & \textbf{69.01} \\
LightSEAD Coarse + DISK & 1,518 & 35.96 & 49.25 & 60.54 \\
Native SuperPoint       & 2,475 & 31.07 & 46.85 & 60.57 \\
Key.Net + DISK          & 3,876.9 & 29.82 & 45.84 & 61.44 \\
Native XFeat            & 4,092 & 28.81 & 45.24 & 60.05 \\
ORB                     & 3,905 & 8.29 & 16.04 & 18.89 \\
\bottomrule
\end{tabular}
\caption{MegaDepth-1500 relative-pose AUC (\%) with a top-4,096 ceiling, MNN
matching, and RANSAC~\cite{fischler1981ransac}. Points/image is the mean retained count.}
\label{tab:megadepth_pose}
\end{table}

\subsection{Efficiency and Deployment}
LightSEAD's six convolutions, 84,289 parameters and 168.6 kB FP16 weights
support efficient inference (Table~\ref{tab:main_native_runtime}). Full and
Coarse share trained weights, varying input resolution rather than architecture.

Across five learned baselines (SuperPoint, Key.Net, DISK, R2D2 and XFeat) and
three datasets, Coarse FP32 reduces latency by 74.92\%, 75.92\% and 70.16\%
on RTX 4070, RTX 4090 and Jetson, respectively. Each value averages the 15
pairwise reductions $100(t_{\mathrm{baseline}}-t_{\mathrm{Coarse}})/t_{\mathrm{baseline}}$;
XFeat remains faster in some comparisons, and timing boundaries follow Section 4.1.

Mixed-precision Coarse acceleration spans 2.24--2.80$\times$, 1.80--2.62$\times$
and 2.52--2.79$\times$ on these devices, respectively. Mixed-precision inference
preserved repeatability, with differences below 0.004 percentage points across
all evaluated GPU platforms. Coarse takes
13.753/7.206/7.761 ms on Jetson H/K/G and reaches 1.100 ms on RTX4090 KITTI06.
At $1226\times370$ KITTI resolution, these correspond to approximately
909 FPS on RTX4090 and 139 FPS on Jetson Orin Nano Super, supporting
high-throughput detection for resource-constrained visual perception.

\subsection{Limitations}

\noindent\textbf{Spatial distribution of detections.}
The primary handcrafted teacher favours textured or structurally salient
regions. LightSEAD may inherit this tendency, concentrating detections in
texture-rich areas and leaving other regions less covered. This may matter for
applications requiring correspondences across a wider field of view, although
persistent features in these regions can provide useful mapping landmarks.

\noindent\textbf{No learned descriptor.}
LightSEAD detects keypoints but has no learned descriptor; feature-matching
applications must pair it with one.

\endgroup
\section{Conclusion}
\label{sec:concl}
LightSEAD achieves more with less by transferring calibrated classical priors,
viewpoint and appearance knowledge through geometrically verified distillation
into a six-convolution, 84,289-parameter student. Teacher complexity remains in
training; competitive repeatability and downstream pose utility coexist with low
detector latency. Mixed-precision LightSEAD Coarse reaches approximately 909 FPS
on RTX 4090 and 139 FPS on Jetson at $1226\times370$. These findings support
structured supervision as a route towards compact visual perception for embedded
and resource-constrained systems, without requiring a larger inference backbone.

{\small
\bibliographystyle{ieeenat_fullname}
\bibliography{main,LightSEAD_Experiments_Results_Compact/references,supplementary/references_extra}
}

\clearpage
\hypersetup{pageanchor=false}
\begingroup
\renewcommand{\thesection}{S\arabic{section}}
\renewcommand{\thefigure}{S\arabic{figure}}
\renewcommand{\thetable}{S\arabic{table}}
\renewcommand{\thepage}{S\arabic{page}}
\renewcommand{\theHsection}{supp.\arabic{section}}
\renewcommand{\theHsubsection}{supp.\arabic{section}.\arabic{subsection}}
\renewcommand{\theHfigure}{supp.\arabic{figure}}
\renewcommand{\theHtable}{supp.\arabic{table}}
\setcounter{section}{0}
\setcounter{figure}{0}
\setcounter{table}{0}
\setcounter{page}{1}
\setlength{\parindent}{1em}
\setlength{\parskip}{3pt}
\captionsetup{font=small,labelfont=bf,skip=4pt,hypcap=false}
\graphicspath{{supplementary/}}
\maketitlesupplementary
\section{Protocol and implementation}
\label{sec:protocol}
Sections~\ref{sec:protocol}--\ref{sec:desc} extend the detector-only analyses in
Sections~4.3--4.5 of the main paper to complete tracking-based relative-pose
frontends. Section~\ref{sec:euroc} presents a sensitivity analysis of
translation-direction estimation using the IMU-informed EuRoC reference
rotation.

LightSEAD is a frozen single-image detector with six convolutions and 84,289
learned parameters. Geometry and the appearance and viewpoint branches
contribute to teacher supervision. LightSEAD Full evaluates the student on the
network input grid. LightSEAD Coarse uses half the resolution in each dimension
and maps selected coordinates back to image pixels. 

\paragraph{Selection and tracking.}
We employ Spatially Balanced Non-Maximum Suppression (SB-NMS), a grid-aware
keypoint selection procedure that combines local response suppression with
spatial allocation. Rather than retaining only the globally highest-confidence
responses, SB-NMS distributes a fixed budget of 500 keypoints across image
regions, allowing lower-response candidates to contribute to downstream
geometric estimation. This procedure operates entirely at inference time and
does not modify or retrain the LightSEAD detector. Pyramidal Lucas--Kanade
tracking (KLT) propagates each source point in the forward direction, then
tracks it backwards to its source image \cite{bouguet2000pyr}. We remove invalid
and out-of-bounds tracks and require a round-trip discrepancy of at most one
pixel. The OpenCV implementation uses a \(21\times21\) window, up to pyramid
level 3, and a termination criterion of 30 iterations or a displacement of 0.01
pixels.

\paragraph{Pose and metric.}
The calibrated five-point estimator uses OpenCV essential-matrix RANSAC followed by cheirality-based recovery \cite{nister2004five}. We use a 2.5-pixel RANSAC threshold, confidence 0.99999 and a cap of 1,000 iterations. We score the larger of the geodesic rotation error and the sign-folded translation direction error
\(\arccos(|\hat{\mathbf t}^{\mathsf T}\mathbf t|/(\|\hat{\mathbf t}\|\|\mathbf t\|))\).
AUC integrates the empirical pose-error success curve up to the stated angular threshold, divided by that threshold. Failed estimates contribute zero success. The metric evaluates relative orientation and direction, with no metric translation scale or trajectory error.

\paragraph{Timing.}
Complete timings start with decoded, preprocessed images and end at pose
recovery. They include device transfers within the timed frontend, selection,
tracking or matching, and geometric fitting. We report milliseconds on an RTX
4070 Laptop GPU and Intel Core Ultra 9 185H host. 

\section{KITTI 06 with shared KLT}
\label{sec:kitti}
We use 1,100 consecutive rectified left-camera pairs from KITTI 06
\cite{geiger2012kitti}. Detectors share forward--backward KLT and pose
estimation while retaining their respective selection policies and budgets.

\begin{center}
\centering
\captionof{table}{KITTI~06 shared-KLT results. Frontend timing is reported as median / p95. Pose AUC values are percentages. Bold denotes the best value for each metric.}
\label{tab:klt}
\scriptsize
\renewcommand{\arraystretch}{1.08}
\setlength{\tabcolsep}{1.2pt}
\begin{tabular*}{\columnwidth}{@{\extracolsep{\fill}}lrrrr@{}}
\toprule
Method/backend & Pts. & KLT & Frontend & AUC @5 / @10 / @20 \\
\midrule
LightSEAD Full PT32 & 500 & 5.44 & 25.77 / 27.86 & 65.74 / 81.45 / 90.43 \\
LightSEAD Full TRT16 & 500 & 5.89 & 15.06 / 17.30  & 65.37 / 81.44 / 90.47 \\
LightSEAD Coarse PT32 & 500 & 5.73 & 13.97 / 16.12 & 62.95 / 80.27 / 89.90 \\
LightSEAD Coarse TRT16 & 500 & 5.52 & \textbf{11.71 / 13.95} & 64.41 / 81.06 / 90.33 \\
\midrule
SuperPoint & 688 & 7.17 & 24.83 / 28.59 & 61.21 / 78.72 / 88.83 \\
XFeat & 4,004 & 33.23 & 57.97 / 63.42 & \textbf{66.06 / 82.14 / 91.01} \\
ORB & 500 & \textbf{4.97} & 17.62 / 20.71 & 25.76 / 43.28 / 59.57 \\
Harris & 636 & 6.10 & 18.16 / 24.18 & 48.35 / 67.74 / 80.96 \\
DISK & 2,713 & 22.49 & 78.85 / 84.89 & 60.77 / 78.35 / 88.69 \\
Key.Net & 1,532 & 15.12 & 143.53 / 150.30 & 64.23 / 80.76 / 90.09 \\
R2D2 & 5,000 & 39.05 & 234.10 / 239.84 & 64.97 / 81.76 / 90.84 \\
 
\bottomrule
\end{tabular*}
\end{center}

As shown in Table~\ref{tab:klt}, LightSEAD Full achieves 65.74 AUC at
\(5^\circ\) with 500 selected keypoints,
compared with 66.06 for XFeat using approximately 4,004 points and 64.97 for
R2D2 using 5,000. Despite the difference in feature budgets, all three
frontends achieve similar pose accuracy on KITTI~06. Reducing LightSEAD's input
resolution in Coarse lowers the FP32 frontend latency from 25.77 to 13.97\,ms,
with a 2.79-point reduction in AUC at \(5^\circ\). This is a useful trade-off
for applications where computational cost matters more than small differences
in pose accuracy.

Table~\ref{tab:kitti_capped_1k} shows that restricting competing detectors to
1,000 points has different effects across methods. XFeat loses 5.46 AUC points
at \(5^\circ\), R2D2 loses 3.12, and DISK
loses 6.78, whereas SuperPoint remains nearly unchanged. LightSEAD Full and
Coarse retain their original 500-point configurations from
Table~\ref{tab:klt}, achieving 65.74 and 62.95, respectively. The results
indicate that the number of selected features
alone is a poor predictor of pose accuracy. 

\newpage
\begin{center}
\captionof{table}{KITTI~06 pose AUC with LightSEAD at 500 points and each competing detector capped at 1,000 selected points. AUC values are percentages; bold denotes the best value at each threshold.}
\label{tab:kitti_capped_1k}
\scriptsize
\renewcommand{\arraystretch}{1.08}
\setlength{\tabcolsep}{2pt}
\begin{tabular*}{\columnwidth}{@{\extracolsep{\fill}}lrr@{}}
\toprule
Method/backend & \shortstack{Mean points\\per frame} & \shortstack{Pose AUC\\@5 / @10 / @20} \\
\midrule
LightSEAD Full PT32 & 500 & \textbf{65.74} / \textbf{81.45} / 90.43 \\
LightSEAD Full TRT16 & 500 & 65.37 / 81.44 / \textbf{90.47} \\
LightSEAD Coarse PT32 & 500 & 62.95 / 80.27 / 89.90 \\
LightSEAD Coarse TRT16 & 500 & 64.41 / 81.06 / 90.33 \\
\midrule
SuperPoint & 687 & 61.27 / 78.75 / 88.85 \\
XFeat & 1,000 & 60.60 / 78.02 / 88.06 \\
ORB & 999 & 34.37 / 52.88 / 68.02 \\
Harris & 999 & 34.91 / 54.01 / 69.22 \\
DISK & 1,000 & 53.99 / 72.81 / 84.87 \\
Key.Net & 1,000 & 58.73 / 76.80 / 87.72 \\
R2D2 & 1,000 & 61.85 / 79.62 / 89.67 \\
 
\bottomrule
\end{tabular*}
\end{center}

\section{Comparison with descriptor frontends}
\label{sec:desc}
Table~\ref{tab:desc} reports descriptor-based variants on the same consecutive
KITTI~06 pairs. SuperPoint, XFeat, DISK and R2D2 use their native learned
descriptors
\cite{detone2018superpoint,potje2024xfeat,tyszkiewicz2020disk,revaud2019r2d2}.
ORB uses its native binary descriptors \cite{rublee2011orb}. For Harris we
extract SIFT descriptors at the detected locations
\cite{harris1988combined,lowe2004sift}. The Key.Net pipeline uses OriNet
orientations and HardNet descriptors
\cite{barrosolaguna2019keynet,mishchuk2017hardnet}. Binary features use Hamming mutual
nearest neighbours (MNN), and the other pipelines use cosine MNN. Pose recovery
uses the same OpenCV configuration.

\begin{center}
\captionof{table}{KITTI~06 descriptor matching. Bold denotes the best value for each metric.}
\label{tab:desc}
\small
\renewcommand{\arraystretch}{1.08}
\setlength{\tabcolsep}{2.4pt}
\begin{tabular*}{\columnwidth}{@{\extracolsep{\fill}}lrrrr@{}}
\toprule
Pipeline & @5 & @10 & @20 & ms \\
\midrule
SuperPoint & 52.21 & 73.04 & 85.88 & \textbf{42.63} \\
XFeat & 48.99 & 71.31 & 85.18 & 42.77 \\
ORB & 17.03 & 34.21 & 53.55 & 44.94 \\
DISK & 58.24 & 77.47 & 88.57 & 108.86 \\
Harris + SIFT & 32.40 & 53.89 & 72.33 & 122.46 \\
R2D2 & 56.85 & 76.97 & 88.47 & 281.72 \\
Key.Net + HardNet & \textbf{62.23} & \textbf{80.07} & \textbf{89.96} & 361.23 \\
 
\bottomrule
\end{tabular*}
\end{center}

\subsection{Correspondence Generation and Frontend Efficiency}
Descriptor-based matching generally incurs higher latency than KLT tracking in
the evaluated pipelines. SuperPoint increases from 24.83\,ms with KLT
(Table~\ref{tab:klt}) to 42.63\,ms with descriptor matching
(Table~\ref{tab:desc}), while DISK increases from 78.85 to 108.86\,ms and
Key.Net from 143.53 to 361.23\,ms. Their pose accuracy is also lower with native
descriptor matching. XFeat is an exception in latency, taking 42.77\,ms with
descriptors compared with 57.97\,ms with KLT, although its AUC at \(5^\circ\)
decreases from 66.06 to 48.99.

LightSEAD avoids descriptor extraction entirely. With KLT, Full and Coarse
achieve 65.37 and 64.41 AUC at \(5^\circ\) at 15.06 and 11.71\,ms,
respectively, using TensorRT FP16. Both are faster than the tested external KLT
frontends in Table~\ref{tab:klt} and the descriptor-based pipelines in
Table~\ref{tab:desc}. These results favor lightweight detection and
optical-flow tracking for consecutive-frame pose estimation.

\section{IMU-informed relative pose estimation on EuRoC}
\label{sec:euroc}
We evaluate on the EuRoC MAV dataset \cite{burri2016euroc}, which consists of
indoor recordings captured by a micro aerial vehicle in motion. We use the
Machine Hall sequences MH\_01\_easy and MH\_03\_medium.

The median displacement between consecutive frames is only 2.23\,cm on
MH\_01\_easy and 4.88\,cm on MH\_03\_medium. At these short baselines, weak
translational parallax can make monocular translation-direction estimates
sensitive to small rotation errors. This motivates evaluating how an
IMU-informed rotation input affects relative pose estimation.

We use 256 evenly spaced anchor pairs per sequence. Each pair spans eight
camera frames, approximately 0.4\,s at the 20\,Hz capture rate; this gap defines
the evaluation interval. We interpolate the released EuRoC reference trajectory
to both camera timestamps and compose it with the body-to-camera calibration to
obtain the relative camera rotation over the same interval as the visual
estimate.

This experiment is a sensitivity analysis of translation-direction estimation.
For each frontend, we hold the correspondences and translation solver fixed
and vary only the rotation input between the visual estimate and the
IMU-informed EuRoC reference rotation. Given rotation \(R\), the fixed
calibrated bearings \(\mathbf b_s,\mathbf b_t\) constrain the unit translation
direction through
\(\mathbf b_t^\mathsf T[\mathbf t]_\times R\mathbf b_s=0\). All frontends use
the same robust hypothesis generation, Sampson threshold, inlier refit, and
sign-folded translation-direction metric.

Figure~\ref{fig:rotation} shows the paired improvement. For LightSEAD
Full, the fraction of pairs with translation-direction error within
\(5^\circ\) rises from 53.52\% to 84.77\% on MH\_01\_easy and from 58.98\% to
89.45\% on MH\_03\_medium. DISK and SuperPoint show similar gains.

Table~\ref{tab:euroc_ref} reports IMU-informed pose AUC and median end-to-end
visual frontend latency. LightSEAD Full TRT16 achieves AUC at \(5^\circ\) of
60.59\% and 63.46\% on MH\_01\_easy and MH\_03\_medium, respectively, with
median latencies of 38.56 and 36.25\,ms. On MH\_03\_medium, DISK achieves
64.09\% AUC at \(5^\circ\) at 105.30\,ms.

\begin{figure*}[t]
\centering
\includegraphics[width=.96\textwidth]{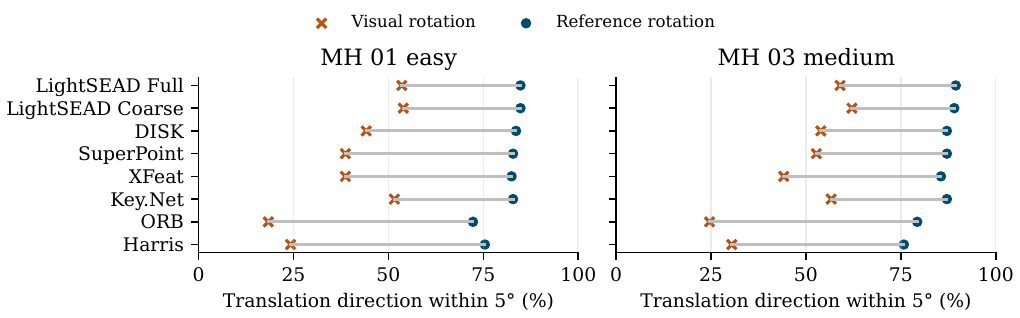}\par
\caption{Paired rotation-input diagnostic at gap eight. Each row uses the same
correspondences for the visual and IMU-informed EuRoC reference rotations.
Circles indicate the IMU-informed reference rotation and crosses indicate the
visual estimate. The cohort contains 256 anchor pairs per sequence.}
\label{fig:rotation}\par\vspace{8pt}
\end{figure*}

\begin{table*}[t]
\centering
\caption{IMU-informed pose AUC at \(5^\circ/10^\circ/20^\circ\) on 256 gap-eight
anchor pairs per sequence. In the visual frontend latency, E denotes
MH\_01\_easy and M denotes MH\_03\_medium; the latency column reports median
end-to-end visual
frontend latency on the directly timed gap-eight visual benchmark. Bold denotes the best value at each threshold and the
lowest latency per sequence.}
\label{tab:euroc_ref}
\small
\renewcommand{\arraystretch}{1.10}
\setlength{\tabcolsep}{4pt}
\begin{tabular*}{\textwidth}{@{\extracolsep{\fill}}lrrr@{}}
\toprule
& \multicolumn{2}{c}{IMU-informed pose AUC (\%)} &
\shortstack{Visual frontend latency\\E/M (ms)} \\
\cmidrule(lr){2-3}
Method/backend & MH\_01\_easy & MH\_03\_medium & \\
\midrule
LightSEAD Full PT32 & \textbf{60.73}/\textbf{73.59}/81.26 & 63.34/77.01/84.54 & 52.73/48.86 \\
LightSEAD Full TRT16 & 60.59/73.53/\textbf{81.29} & 63.46 / \textbf{77.07} / 84.57 & 38.56/36.25 \\
LightSEAD Coarse PT32 & 59.66/72.97/80.58 & 62.83/76.64/84.52 & 33.42/\textbf{32.94} \\
LightSEAD Coarse TRT16 & 59.77 / 73.27 / 80.87 & 62.85 / 76.81 / \textbf{84.63} & \textbf{32.86}/34.79 \\
\midrule
DISK & 58.91/72.10/79.48 & \textbf{64.09}/76.75/84.09 & 107.56/105.30 \\
SuperPoint & 57.94/71.45/79.51 & 61.87/75.61/83.58 & 53.81/53.84 \\
XFeat & 55.39/69.93/78.58 & 60.13/74.13/82.62 & 57.15/57.28 \\
Key.Net & 58.09/71.52/79.18 & 62.63/75.80/83.44 & 323.79/314.79 \\
ORB & 42.85/61.33/73.48 & 50.87/66.85/76.69 & 50.71/53.76 \\
Harris & 46.76/63.69/75.21 & 49.82/65.00/76.28 & 127.61/111.82 \\
 
\bottomrule
\end{tabular*}
\end{table*}

The timings use C++ OpenCV 4.5.4/TBB. LightSEAD latency includes each complete
eight-step frontend through pose recovery; external latency includes descriptor
latency and pose recovery.

\section{Remarks}
LightSEAD demonstrates how selective knowledge distillation can transfer
classical geometric priors, viewpoint, and appearance knowledge into a compact
six-convolution detector with 84,289 parameters. Beyond detection, LightSEAD
supports low-latency, descriptor-free tracking with competitive relative pose
accuracy on KITTI 06 and EuRoC MAV under the evaluated protocols. Furthermore,
on EuRoC MAV, using the IMU-informed reference rotation improved
translation-direction accuracy with image correspondences held fixed.

\endgroup
\end{document}